\documentclass[journal]{IEEEtran}
\usepackage{generic}
\usepackage{cite}
\usepackage{amsmath,amssymb,amsfonts}
\usepackage{algorithmic}
\usepackage{graphicx}
\usepackage{algorithm,algorithmic}
\usepackage{textcomp}
\usepackage{bm}
\usepackage{multirow}
\usepackage{subfigure}
\usepackage{diagbox}
\usepackage{mathtools}
\usepackage{booktabs} 
\usepackage{ulem}
\usepackage{url}
\usepackage{xcolor}
\usepackage[breaklinks]{hyperref}
\hypersetup{colorlinks=true,
            linkcolor=blue,
            anchorcolor=blue,
            citecolor=blue}

\def\BibTeX{{\rm B\kern-.05em{\sc i\kern-.025em b}\kern-.08em
    T\kern-.1667em\lower.7ex\hbox{E}\kern-.125emX}}
\begin{document}
\bstctlcite{IEEEexample:BSTcontrol}

\title{CAMANet: Class Activation Map Guided Attention Network for Radiology Report Generation}
\author{Jun Wang, Abhir Bhalerao, Terry Yin, Simon See, \IEEEmembership{Senior Member, IEEE}, and Yulan He
\thanks{Jun Wang's PhD studentship is jointly funded by the University of Warwick and China Scholarship Council. We would like to thank the University of Warwick's Scientific Computing Group for their support of this work and 
the UKRI/EPSRC HPC platform, Sulis and Avon. }
\thanks{Jun Wang and Abhir Bhalerao are with the Department of Computer Science, University of Warwick, CV4 7AL Coventry, UK. (email: jun.wang.3@warwick.ac.uk, abhir.bhalerao@warwick.ac.uk)
}
\thanks{Terry Yin and Simon See are with the NVIDIA AI Tech Center, Singapore.}
\thanks{Yulan He is with the Department of Informatics, King’s College London, WC2R 2LS, London, UK, the Department of Computer Science, University of Warwick, CV4 7AL Coventry, UK and the Alan Turing Institute, UK.}
}

\maketitle

\begin{abstract}
Radiology report generation (RRG) has gained increasing research attention because of its huge potential to mitigate  medical resource shortages and aid the process of disease decision making by radiologists. Recent advancements in RRG are largely driven by improving a model's capabilities in encoding single-modal feature representations, while few studies explicitly explore the cross-modal alignment between image regions and words. Radiologists typically focus first on abnormal image regions before composing the corresponding text descriptions, thus cross-modal alignment is of great importance to learn a RRG model which is aware of abnormalities in the image. Motivated by this, we propose a \textbf{C}lass \textbf{A}ctivation \textbf{M}ap guided \textbf{A}ttention \textbf{Net}work (CAMANet) which explicitly promotes cross-modal alignment by employing aggregated class activation maps to supervise cross-modal attention learning, and simultaneously enrich the discriminative information. 
CAMANet contains three complementary modules: a Visual Discriminative Map Generation module to generate the importance/contribution of each visual token; Visual Discriminative Map Assisted Encoder to learn the discriminative representation and enrich the discriminative information; and a Visual Textual Attention Consistency module to ensure the attention consistency between the visual and textual tokens, to achieve the cross-modal alignment. Experimental results demonstrate that CAMANet outperforms previous SOTA methods on two commonly used RRG benchmarks.~\footnote{The code is available at \hyperlink{here}{https://github.com/Markin-Wang/CAMANet}}
\end{abstract}

\begin{IEEEkeywords}
Radiology report generation, Cross-modal alignment, Class activation map
\end{IEEEkeywords}

\section{Introduction}
\label{sec:introduction}
Radiology report generation (RRG) aims to automatically describes radiology images, e.g., X-Ray and MRI, by human-like language. Generating a coherent report requires expertise from radiologists, who are however among the most in-demand medical specialists in most countries. Moreover, it can take at least 5 minutes to describe a radiology image even for a professional radiologist 
. Consequently, there had been growing interest in automating RRG because of its huge potential to efficiently and effectively assist the diagnosis process.

\begin{figure}
    \centering 
    \includegraphics[width=0.45\textwidth]{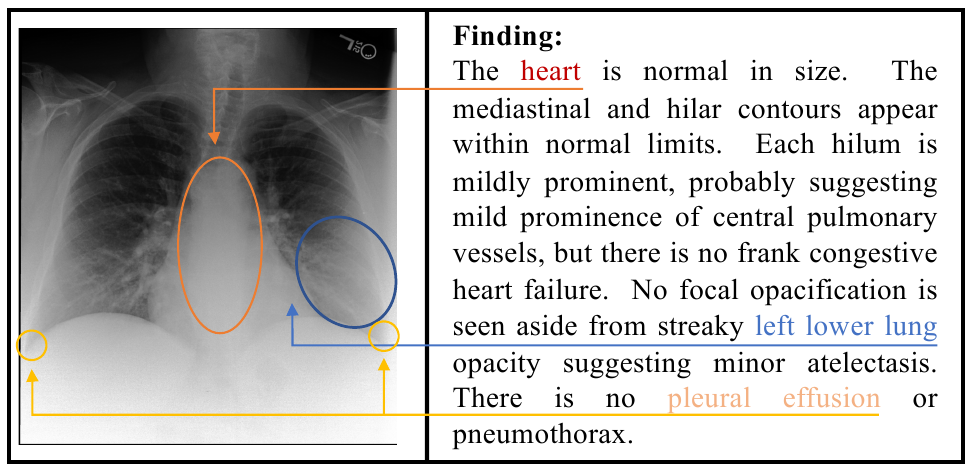} 
    \caption{A Chest XRay image with its report findings. By way of example, manually aligned visual-textual features are marked in the same color}.   
    \label{fig:motivation}  
    \vspace{-20pt}
\end{figure}

RRG is a challenging task 
but with the availability of large-scale datasets and newly developed high-performance computer vision and language models, some valuable insights and improvements have been recently reported~\cite{chen2020generating,chen2021cross,wang2022cross,liu2019clinically}. The performance however is still far from satisfactory for methods to be deployed in clinical practice. This is because, different from the traditional image captioning task which mostly requires generating one or two sentences, RRG demands $2$-$4$ times \textit{more} text to adequately describe the findings in an image. Also, radiology reports tend to contain sentences with more sophisticated semantic relationships with their corresponding image regions, calling for a need for \textit{more precise} cross-modal alignment, i.e., alignment between words and image regions. An example is shown in \autoref{fig:motivation} where the cross-modal alignments are marked in the same color. The problem is exacerbated as there exist data biases in commonly used datasets with significantly fewer radiology reports from X-ray images containing abnormal regions, making it hard for the RRG models to efficiently capture 
abnormal features. Also, such abnormal regions often make up only a small proportion of pathological images. Furthermore, even in pathological cases, most report sentences may be associated with a description of normal findings, as shown in \autoref{fig:motivation}.

Previous approaches~\cite{zhang2020radiology,chen2020generating,chen2021cross} often focus on improving the visual representation capability or better learning the semantic patterns by utilizing a learnable memory. Few studies explicitly explore representation learning with information about image abnormality detection and cross-modal alignment. To bridge this gap, we propose a Class Activation Map guided Attention Network (CAMANet) which distills the discriminative information into the encoder and explicitly improves cross-modal alignment by leveraging class activation maps~\cite{zhou2016learning}. 
This work makes four principal contributions:
\begin{enumerate}
\item We propose a novel end-to-end class activation map guided attention model where the class-activation map (CAM) is utilized to explicitly promote cross-modal alignment and discriminative representation learning. To the best of our knowledge, CAMANet is the first work to leverage CAMs in this way.

\item A Visual Discriminative Map (VDM) Generation module to derive the visual discriminative map from the CAMs based on the pseudo labels from an automatic labeler.

\item We present a VDM Assisted Encoder which enriches the discriminative information by utilizing a self-attention mechanism and the VDM.

\item We design a visual-textual attention consistency module which considers the VDM as the ground truth to supervise the cross-modal attention learning in the decoder, promoting cross-modal alignment.
\end{enumerate}

Experimental results demonstrate that CAMANet outperforms the previous state-of-the-art (SOTA) methods on two widely-used benchmarks. 
Discussion and proposals are given to inspire future work. 

\begin{figure*}
    \centering 
    \includegraphics[width=0.95\textwidth]{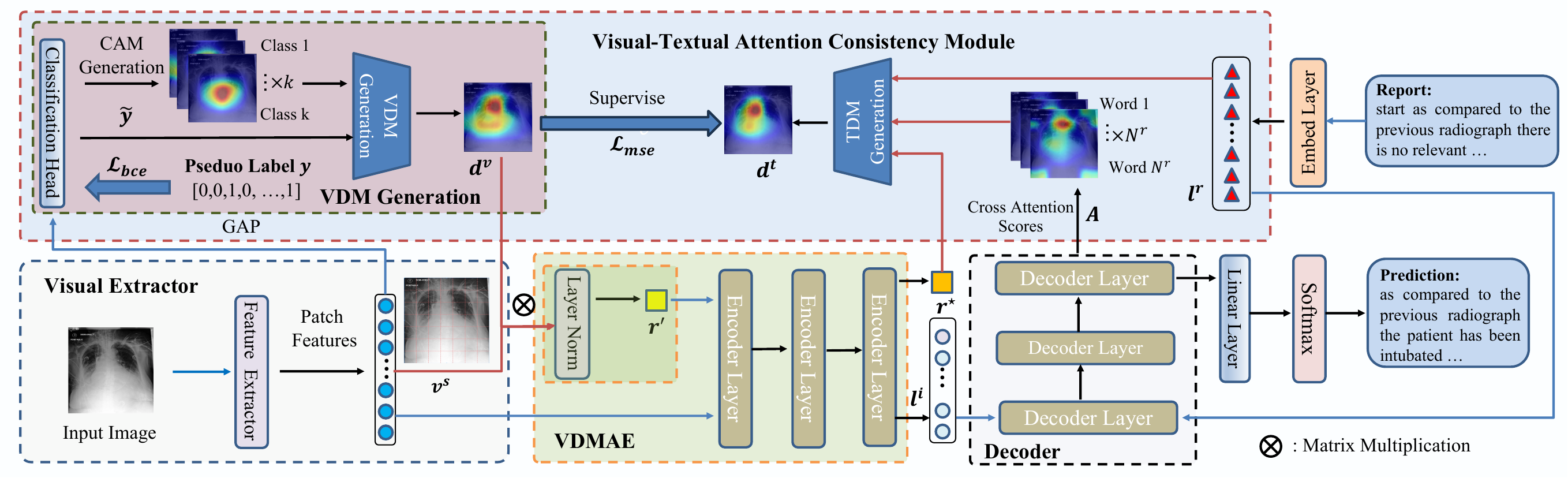} 
    \caption{The architecture of CAMANet: An image is fed into the Visual Extractor to obtain patch features which are then utilized to generate the VDM via the VDM Generation module. The proposed VDMAE leverages the VDM to derive a discriminative representation and enrich the discriminative information. Combined with word embedding, visual tokens are sent to the transformer to produce the report. After that, the VTAC module generates a TDM from the cross-modal attention scores and considers the VDM as the ground truth to supervise the cross-modal alignment learning. }    
    \label{fig:architecture}  
    \vspace{-10pt}
\end{figure*}

\section{Related Work}
\label{sec:related_works}
\subsection{Image Captioning}
Image captioning, aiming to describe an image with human-like sentences, is considered as a high-level visual understanding problem taking advantage of both the techniques of computer vision and natural language processing. Because of the great success in machine translation and language generation, recent SOTA approaches~\cite{anderson2018bottom,guo2020normalized, pan2020x, rennie2017self,lu2017knowing} also follow an encoder-decoder architecture and have demonstrated a great improvement in some traditional image captioning benchmarks. In particular, the most successful models~\cite{cornia2020meshed,guo2020normalized,pan2020x,ji2021improving} usually adopt the Transformer~\cite{vaswani2017attention} as their backbone due to its self-attention mechanism and impressive capability of modelling long-range dependency. However, these methods are tailored for traditional scene images and not suitable for the radiology images with long sentence report and fine-grained, abnormal regions. Although several works~\cite{krause2017hierarchical,melas2018training} have been developed to tackle long text generation, they often fail to capture specific medical observations and tend to generate normal descriptions, resulting in unsatisfactory performance. 

\subsection{Radiology Report Generation}
\label{ssec:rrg}
Following on from the success in image captioning, recent SOTA RRG studies all adopt similar architectures: an encoder-decoder to generate the report combined with a visual feature extractor. For example, Jing et al.\cite{jing2018automatic} proposed a hierarchical LSTM model to mitigate the long sentence generation problem and developed a co-attention mechanism to localize abnormal regions. Other works have utilized disease topic information~\cite{liu2019clinically, zhang2020radiology} and obtain better results. Liu et al.~\cite{liu2021exploring} extended these works by taking advantage of both the prior and posterior knowledge, i.e., knowledge from the similar images and a predefined topic bag. Chen et al.~\cite{chen2020generating} designed a relational memory and a memory-driven conditional layer normalization to better learn the report patterns.

The aforementioned studies often neglect the importance of cross-modal alignment and discriminative (disease-related) representation learning. As the only work we know exploring the cross-modal alignment in RRG, R2GenCMN~\cite{chen2021cross}, designed a shared memory matrix, expected to be able to implicitly learn the cross-modal alignment. However, learning the cross-modal alignment in this way proves difficult since the model is only trained under a cross entropy loss for report generation without any other forms of supervision/guidance for the cross-modal alignment learning. To address these problems, we propose a visual-textual attention consistency module to explicitly force the visual and textual modalities to focus on the same image regions. In addition, we develop a VDM-assisted encoder to enrich the discriminative information.

There are a few works~\cite{zhou2020more, chen2018knowledge} exploring the image-word matching in scene images, which are similar to the cross modal alignment. However, these methods normally require a well pre-trained fully-supervised detector, e.g 
Faster RCNN~\cite{ren2015faster} trained on ground truth labels to detect the object proposals offline. In addition, they exploit bounding-box annotations and utilize them to form the mapping during supervision. Unfortunately, because they are time-consuming and costly to obtain, there is no large dataset for the RRG with such valuable ground truth labels to train an accurate detector and form the extra supervision. As a result, object detection methods become unsuitable for RRG, and instead we seek to integrate and train the visual extractor in the entire architecture. Another significant difference between our work and~\cite{zhou2020more} is that the supervision of~\cite{zhou2020more} is applied on each word-region pair thus making it difficult to learn, while our method improves the cross-modal alignment by ensuring the consistency between the block-level, derived visual and textual discriminative maps. 

\vspace{-5pt}
\subsection{Class Activation Maps}

First proposed by Zhou et al.~\cite{zhou2016learning}, a class activation map (CAM) was widely used for weakly supervised object detection (WSOD) aiming to localize objects with only image-level annotations. Bae et al.~\cite{bae2020rethinking} improved the vanilla CAM method in WSOD by dealing with negatively weighted actions and instability in the thresholding process. Jiang et al.~\cite{jiang2021layercam} observed that previous CAM methods were inclined to locate coarse-grained objects due to the low spatial resolution in the final convolution layer and proposed a LayerCAM to collect object information from coarse to fine levels. Xie et al.~\cite{xie2021online} argued that previous works in WSOD generated a CAM based on high-level features but a CAM from low-level features was important also to include richer contextual object information. Generally, the CAM technique is used mainly to localize the class-specific image regions which are then utilized to obtain the bounding boxes in WSOD~. 

CAMs can identify the discriminative regions in an image by employing a \textit{global average pooling layer} (GAP). Consequently, this technique is widely adopted in some weakly supervised semantic segmentation (WSSS) methods to generate pseudo-labels and provides performance gains. For instance, Sun et al~\cite{sun2021ecs} proposed a ECS-Net which 
regards the derived CAM as the segmentation supervision to drive the model to learn a robust representation. Chen et al.~\cite{chen2022class} pointed out the inconsistency between the pseudo masks generated by the CAM and the binary-cross-entropy loss in WSSS and designed the ReCAM module to cope with this problem. Ru et al.~\cite{ru2022learning} combined the studies of CAMs and Transformers to refine the initial pseudo-labels generated by a CAM via an affinity from an attention module, showing promising results in WSSS.

Different from these works, here we use a CAM to enhance the discriminative representation capability and the cross-modal alignment. Specifically, we utilize the CAM technique to derive a visual discriminative map focusing on the discriminative (important or abnormal) regions which is then used to enrich the discriminative information in the encoder. This acts as the ground truth to supervise the cross-modal attention learning. To the best of our knowledge, we have not seen CAMs being leveraged in this way for the RRG task.

\subsection{Pretrained Language and Vision-Language Models in RRG } Recently, several large language models (LLMs) \cite{liu2023radiology,singhal2023towards,boecking2022making} have been proposed for the radiology domain. However, they are single-modal (language) models, e.g., Radiology-GPT \cite{liu2023radiology}, focusing on text-to-text or text-to-choice tasks, e.g, from radiology findings to radiology impressions and do not support images as input, while RRG aims to generate the findings from an image and is a cross-modal task, with no text available during the inference. Therefore, these LLMs cannot normally be applied directly to RRG.

Although not directly applicable, to still take advantage of the power of pre-trained language models (PLMs), the recent work PromptRRG \cite{wang2023can} leverages PLMs as a separate text encoder with prompt learning to distill prior knowledge into the model for RRG. Pre-trained vision-language models (PVLMs) demonstrate improved performance when applied to the downstream tasks, e.g., image classification and visual-question answering, in the natural scene domain. There are several attempts \cite{moon2022multi,delbrouck2022vilmedic} exploring the performance of PVLMs in RRG. However, although they leverage extra datasets and utilize both the findings and impressions sections of reports, these methods have shown unpromising results in RRG possibly since the simple pre-training scheme cannot cope with the intrinsic problems in RRG, aforementioned, and inadequate pretraining on paired medical data for a cross-modal task such as RRG.

\section{Method}
Before detailing our method, in Section~\ref{sec:background} we provide some background of RRG including problem formulation and general architecture. In Section~\ref{sec:CAMANet}, we elaborate each proposed component of CAMANet. The final objective function is  described in Section~\ref{sec:objective}. An overview of the CAMANet architecture is illustrated in~\autoref{fig:architecture}. Generally, an image is first fed into the visual feature extractor to obtain a sequence of visual tokens. Then, the proposed visual discriminative map (VDM) generation module takes these visual tokens as input to derive the visual discriminative map focusing on possible abnormal regions. This VDM, combined with the visual tokens, are then sent to the VDM assisted encoder module to enrich the abnormal information and then generate the report in the decoder. The VTAC module generates a textual discriminative map (TDM) from the cross-modal attention scores and considers the VDM as the ground truth to supervise cross-modal alignment learning.

\subsection{Background}
\label{sec:background}

Given a radiology image $\bm{I}$, 
the purpose of RRG is to generate coherent findings (report) $\bm{R}$ from $\bm{I}$. Recent SOTA methods often adopt an encoder-decoder framework to generate the report. In particular, a visual extractor, e.g. DenseNet121~\cite{huang2017densely}, is first employed to extract visual features $\bm{V}^i \in \mathbb{R}^{H \times W \times C}$ that are then flattened to a sequence of visual tokens $\bm{V}^s \in \mathbb{R}^{HW \times C}$.  $H$, $W$, $C$ are the height, width and the number of channels respectively. This process is formulated as:
\begin{equation}
\label{eq:vt}
 \{ v^s_1, v^s_2, ...,v^s_k, ..., v^s_{N^s-1}, v^s_{N^s} \} = f_{vfe}(\bm{I}),
\end{equation}
\noindent
where $v^s_k$ denotes the patch feature in the $k^{th}$ position in $\bm{V}^s$, and $N^s=H\times W$. $f_{vfe}$ is the visual feature extractor.  

These visual tokens are then fed into the transformer based encoder-decoder to generate a report $\bm{R}$. Specifically, at time step $T$, the encoder maps the visual tokens into an intermediate representations $\bm{l}^i \in \mathbb{R}^{1\times D}$. An embedding layer is applied to obtain the word embedding $\bm{l}^r \in \mathbb{R}^{1\times D}$ of each word $w$ in the report $\bm{R}$. $D$ is the number of dimensions of hidden states. After this, the decoder takes these two sequence features as source inputs and predicts the current output (word). In general, we express the encoding and decoding processes as: 
\begin{align}
\label{eq:encoder}
 \{l^i_1, l^i_2, ...,  l^i_{N^s}\} = f_{en}(v^s_1, v^s_2,..., v^s_{N^s}), \\
  \{l^r_1, l^r_2, ...,  l^r_{T-1}\} = f_{em}(w_1, w_2,..., w_{T-1}), \\
\label{eq:decoder}
  p_T = f_{de}(l^i_1, l^i_2, ...,  l^i_{N^s};l^r_1, l^r_2, ..., l^r_{T-1}),
\end{align}

\noindent
Where the $f_{en}, f_{em}$ and $f_{de}$ denotes the encoder, embedding layer and the decoder respectively. $w_i$ and $p_T$ are the $i^{th}$ word in the report and the word prediction at time step $T$.

The encoder and decoder both consist of several transformer layers where a cross-attention is further added after the self-attention in each decoder layer to enable cross-modal fusion.

\subsection{CAMANet}

\label{sec:CAMANet}
Learning the cross-modal alignment and discriminative representation is challenging but essential for RRG. To achieve this, CAMANet uses a Visual Discriminative Map Assisted Encoder and a Visual-Textual Attention Consistency module which take advantage of the CAM technique.

\subsubsection{Visual Discriminative Map Generation}

The first step is to generate a visual discriminative map indicating the discriminative regions in an image. This  map is then used in the VDM Assisted Encoder (VDMAE) and the Visual-Textual Attention Consistency (VTAC) modules to enable the discriminative representation learning and cross-modal alignment. \par

We propose a way to leverage the class activation map (CAM) technique to localize discriminative regions. CAM requires the image category labels to form a classification task, so as to obtain the patch contributions (importance) to each category. The category labels are often unavailable for RRG datasets because of the time and expense required to create them. To overcome this problem in CAMANet, we utilize CheXpert~\cite{irvin2019chexpert}, an automatic labeller, to generate a pseudo label $\bm{y}$ for each image. We can form a multi-label classification task given that $\bm{y}$ is a multi-hot vector encoding the presence of 14 common observations in Chest X-ray images from the automatic labeller.
As shown in the upper lef red box in ~\autoref{fig:architecture}, a classification head is added, taking the extracted visual tokens from the visual extractor as input, to predict the presence $\widetilde{y}_i \in {0,1}$ for $i^{th}$ disease formulated as:
\begin{align}
\label{eq:cls}
\bm{v}_g =& \frac{1}{N^s}\sum_{j=1}^{N^s}v^s_j, 
\\
\{\widetilde{y}_1, \widetilde{y}_2,...,\widetilde{y}_{N^c}\} =& \       \Phi(\text{Softmax}(\bm{W}_c \cdot \bm{v}_g)),
\\
\Phi(o) =& \left\{
\begin{aligned}
0 \quad    if \ o \leq  0.5, \\
1 \quad    if \ o > 0.5, \\
\end{aligned}
\right.
\end{align}
where $\bm{v}_g$ and $\bm{W}_c$ are the global visual feature and weights of the classification head. $N^c$ is the number of disease categories. 

After the forward process of the visual extractor, we can obtain the class activation map $\bm{m}^i=\{m^i_1, m^i_2,...,m^i_{N^s}\}$ for $i^{th}$ category with $\widetilde{y}_i=1$. Remember that the classification is for the multi-label scenario, hence each sample may show the presence of multiple observations. To this end, we aggregate the class activation maps for all the categories present and generate a visual discriminative map $\bm{d}^v$ to target important regions. In particular, we first use the ReLU activation function to zero out negative contributions. Min-max normalization is then applied to each class activation map. \autoref{eq:norm} formulates this process on the class activation map of $i^{th}$ category $\bm{m}^i$:
\begin{equation}
\label{eq:norm}
\widetilde{m}^i_{j} = \frac{\sigma(m^i_j)-\min(\sigma(\bm{m}^i))}{\sigma(\max(\bm{m}^i))-\sigma(\min(\bm{m}^i))},
\end{equation}
where $\widetilde{m}^i_{j}$ is the normalized contribution of $j^{th}$ patch for $i^{th}$ category and $\sigma$ denotes the ReLu activation function. Then, the final visual discriminative map $\bm{d}^v \in \mathbb{R}^{1 \times N^s}$ is obtained via maximum pooling over all the class activation maps:
\begin{equation}
\label{eq:maxpool}
\bm{d}^v = \text{MaxPool}(\{\bm{m^k}|\, \widetilde{y}_k=1\}).
\end{equation}

Note that we chose the class activation map techniques to generate the visual discriminative map due to their high efficiency, and any weakly supervised techniques producing patch-level scores could be applied here. 

\subsubsection{VDM Assisted Encoder}
\label{sec:vdm}
Since the dataset is dominated by normal samples, e.g., normal images and descriptions, the RRG model struggles to capture  abnormal information and describe abnormal regions. To mitigate this problem, we propose a visual discriminative map (VDM) assisted encoder, the VDMAE, to enrich the discriminative information. This is achieved by learning a discriminative representation in terms of the VDM $\bm{d}^v$, which is distilled into the encoder. We show this procedure by the green rectangle in \autoref{fig:architecture}. Specifically, the discriminative representation $\bm{r}^{'} \in \mathbb{R}^{1\times C}$ is obtained by matrix multiplication between the visual discriminative map $\bm{d}^v \in \mathbb{R}^{1 \times N^s}$ and the image patch features followed by a normalization layer to stabilize the distribution,
\begin{align}
\label{eq:dr}
\bm{r} &= \bm{d}^v\bm{V}^s, \\
\label{eq:layernorm}
\bm{r}^{'} &= \text{LayerNorm}(\bm{r}).
\end{align}

where $\bm{V}^s \in \mathbb{R}^{N^s\times C}$ is the extracted visual token sequence obtained by \autoref{eq:vt}.

After obtaining the discriminative representation, 
we add it as an extra 
input token to the encoder and leverage the power of an attention mechanism to gradually incorporate the VDM features for representation learning. Then~\autoref{eq:encoder} becomes:
\begin{equation}
\label{eq:new_en}
\{\bm{r}^{\star}, l^i_1, ...,  l^i_{N^s}\} = f_{en}(\bm{r}^{'}, v^s_1, ..., v^s_{N^s}),
\end{equation}
where $\bm{r}^{\star}$ denotes the encoded discriminative representation. Through the self-attention mechanism in the encoder, other visual tokens can fully interact with this to learn the useful discriminative information. 


\begin{figure}
    \centering 
    \includegraphics[width=0.4\textwidth]{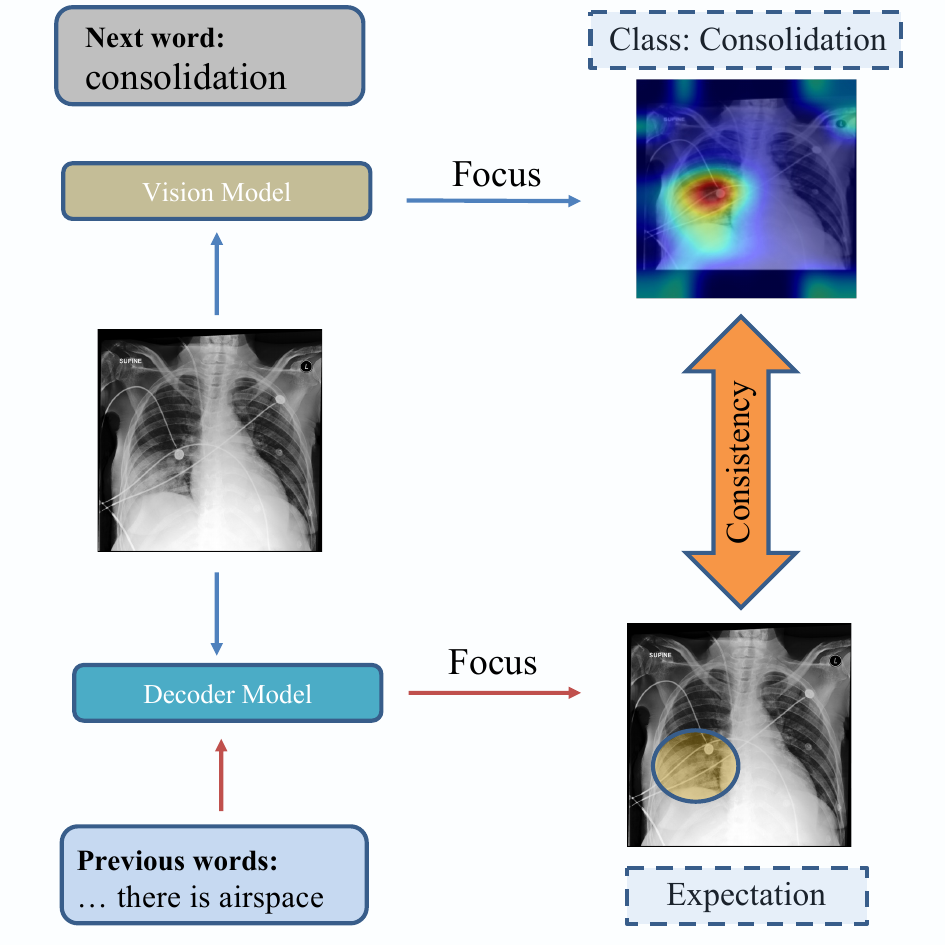} 
    \caption{An illustration of the visual-textual attention consistency. 
    The decoder model is expected to pay attention to the same regions as the vision model to achieve consistency.}    
    \label{fig:example_con}  
    \vspace{-10pt}
\end{figure}

\subsubsection{Visual-Textual Attention Consistency}
\label{sec:vtac}
Human experts demonstrate excellent consistency when generating the report, i.e., they pay attention to important regions first and are able to accurately describe what they see. To exploit such characteristics, we design a Visual-Textual Attention Consistency (VTAC) module to explicitly learn the cross-modal alignment using the visual discriminative map $\bm{d}^v$ as the ground truth to supervise the cross-modal attention learning in the decoder, as demonstrated in the upper right blue rectangular in \autoref{fig:architecture}. When generating a finding, like human behaviour, we expect that the decoder model should attend to the same regions as the visual classifier which is trained to focus on the important regions described in the previous sections. \autoref{fig:example_con} shows an illustration of this visual-textual attention consistency. In this example, the VDMAE module will produce an attention map focusing on the relevant regions for a sample presenting ``Consolidation", and the decoder model is expected to pay attention to the same regions as the vision side in the cross-modal attention when generating the next word ``consolidation" given the previous words and visual tokens. 



Nonetheless, each word in the report has its own attention to the image in the cross-modal attention. This raises a question of how to utilize a single \textit{visual} discriminative map to supervise multiple attentions from different words. To deal with this problem, we also form a textual discriminative map $\bm{d}^t$ by aggregating the attentions from  \textit{important} words. These important words are selected based on the similarities between their word embeddings and encoded visual discriminative representation $\bm{r}^\star$ which contains rich visual discriminative information. In detail, we first compute the cosine similarities between the embedding of each word and the visual discriminative representation via:
\begin{equation}
\label{eq:sim}
s_j = \frac{l^r_j \cdot \bm{r}^{\star}}{|l^r_j|| \bm{r}^{\star}|},
\end{equation}
where $s_j$ denotes the similarity between $j^{th}$ word and the discriminative representative and $\cdot$ is the dot product operation.

We expect the embeddings of important words to have higher similarity to the visual discriminative representation in the latent space. Therefore, the top $k\%$ words with the higher similarities are selected to generate a textual discriminative map. 
Specifically, taking the visual tokens and textual tokens representations from the last layer, the cross-modal attention layer in the last decoder outputs the attention scores $A^\star \in \mathbb{R}^{N^r \times N^s} $ between each visual-textual-token pairs. $N^r$ denotes the number of words in the report. After selecting top $k$\% important words through 
\autoref{eq:sim}, we can now form the cross-modal attention score matrix of the important words $\bm{A}=\{\bm{a}^1,\bm{a}^2,...,\bm{a}^{\gamma}\}$ from the whole cross-modal attention scores $\bm{A}^\star$, where $\bm{a}^i \in \mathbb{R}^{1\times N^s}$ denotes the attention scores (similarities) between $i^{th}$ selected important word and each visual tokens (image regions). 
Then, the generation of the \textit{textual} discriminative map $\bm{d}^t \in \mathbb{R}^{1\times N^s}$ is derived via ~\autoref{eq:a_norm} and ~\autoref{eq:a_maxpool}, which is similar to the generation of the \textit{visual} discriminative map,
\begin{align}
\label{eq:a_norm}
\widetilde{a}^i_j =& \frac{\sigma(a^i_j)-\min(\sigma(\bm{a}^i))}{\sigma(\max(\bm{a}^i))-\sigma(\min(\bm{a}^i))}
\\
\label{eq:a_maxpool}
\bm{d}^t =& \text{MaxPool}([\widetilde{\bm{a}}^1,\widetilde{\bm{a}}^2,...,\widetilde{\bm{a}}^{\gamma}]),
\end{align}
where $\widetilde{a}^i_{j}$ is the normalized attention score of $i^{th}$ word for $j^{th}$ image region. $\gamma=\lceil k\%\times N^r \rceil$ is the number of selected words to generate the textual discriminative map. 

However, since~\autoref{eq:a_maxpool} 
ignores the relative importance of the selected words, we use the calculated similarity to generate the weights of each word via $w_j=\sigma(s_j)$. By considering the relative importance, \autoref{eq:a_maxpool} then becomes:
\begin{equation}
\label{eq:w_maxpool}
\bm{d}^t = \text{MaxPool}([w_1\widetilde{\bm{a}^1},w_2\widetilde{\bm{a}^2},...,w_{\gamma}\widetilde{\bm{a}^{\gamma}}]).
\end{equation}
After obtaining the textual discriminative map, $\bm{d}^t  \in \mathbb{R}^{1\times N^s}$, the visual-textual attention consistency is achieved by making $\bm{d}^t$ to be close to the \textit{visual} discriminative map $\bm{d}^v$ 
through a Mean Squared Error (MSE) loss:
\begin{equation}
\label{eq:mse}
 \mathcal{L}_{mse} =\frac{1}{N^s}\sum_{j=1}^{N^s}||\bm{d}^t_j-\bm{d}^v_j||,
\end{equation}
\par
Note that we detach the encoded visual discriminative representation $\bm{r}^\star$ in the decoder. The reason is that adding it as an extra visual token into the decoder will result in a size difference between the visual discriminative map and textual discriminative map ($N^s$ {\em vs} $N^s+1$), since there are no discriminative tokens available before generating the VDM. As mentioned in the Introduction, commonly used datasets suffer from severe data biases. 
The VTAC module therefore aims to explicitly promote the cross-modal alignments by ensuring the language model focusing on the same discriminative visual regions as the vision backbone. The vision and language backbone are complementary eventually achieving consistency to pay greater attention to the same discriminative regions, instead of only relying on cross-modal attention in the decoder in the base model which exhibit almost no cross-modal alignment (we show this in the next section). By enforcing the model to focus more on the discriminative regions, the model can better capture the abnormal features and descriptions, so as to alleviate the data bias. Additionally, 
to explicitly improve the cross-modal alignment in the VTAC module, the model is required to learn how to select the important words whose cross-modal attention scores will be aggregated to generate the textual discriminative map. This encourages the model gain a better understanding of the global context, e.g., sentence discourse relations, of a report.

\subsection{Objective Function}
\label{sec:objective}
Given the entire predicted word token sequence $\{ p_i \}$ as the generated report and the associated ground truth report $\{ w_i \}$, CAMANet is jointly optimized with the cross-entropy loss $\mathcal{L}_{ce}$, the binary cross entropy loss $\mathcal{L}_{bce}$ for the multi-label classification in visual discriminative map generation and the mean square error loss $\mathcal{L}_{mse}$ in the VTAC module:
\begin{align}
\label{final_loss}
 \mathcal{L}_{ce} &= -\frac{1}{N^r}\sum_{i=1}^{N^r}w_i\cdot log(p_i), \\
\mathcal{L} &= \mathcal{L}_{ce} +\lambda \mathcal{L}_{bce} + \delta \mathcal{L}_{mse}.
\end{align}
Here $\lambda$ an $\delta$ are two hyper-parameters which balance the loss contributions.

\section{Experiments}
\label{sec-experiments}

\subsection{Datasets}
We validate the effectiveness of CAMANet on two commonly used RRG datasets, i.e., IU-Xray and MIMIC-CXR. 
IU-Xray
contains 7,470 X-ray images and 3,955 corresponding reports. The majority of patients provided both the frontal and lateral radiology images. MIMIC-CXR
is a large chest X-ray dataset with 473,057 X-ray images and 206,563 reports 
Both of these two datasets are publicly available. We follow the same data splits as~\cite{chen2020generating,chen2021cross}, to divide the IU-Xray dataset into train (70\%), validation (10\%) and test (20\%) sets and remove samples without both view of images. The official data split is adopted for the MIMIC-CXR dataset.

\subsection{Evaluation Metrics}
Four widely used text generation evaluation metrics: BLEU\{1-4\}~\cite{papineni2002bleu}, Rouge-L~\cite{lin2004rouge}, METEOR~\cite{denkowski2011meteor} and CIDEr~\cite{vedantam2015cider} are employed to gauge the model performance. \textbf{BLEU} score is assessed based on n-gram precision. It calculates the overlap of n-grams between the generated and reference texts with a brevity penalty to prevent very short generated sentences. \textbf{Rouge-L} uses the longest common subsequence (LCS) between the generated and reference texts to calculate a score. \textbf{METEOR} takes the synonyms and stemming into the consideration when calculating the score and combines precision and recall into a single metric, giving more weight to recall. \textbf{CIDEr} is a consensus-based scorer. It calculates the term frequency-inverse document frequency (TF-IDF) for n-grams in the generated and ground truth captions to capture the consensus and penalizes common n-grams that are not informative. To measure the model capability of capturing the abnormalities, we follow previous works to report the clinical efficacy metric where CheXbert \cite{smit2020combining} (a recent SOTA automatic labeler supporting the GPU acceleration) is applied to labeling the generated reports and the results are compared with ground truths in 14 different categories related to thoracic diseases and support devices. We use the micro-average precision, recall and F1 score to model the model performance for the clinical efficacy.

\begin{table*}[ht]
\caption{Comparative results of CAMANet with previous studies. The best values are highlighted in bold and the second best are underlined. BL and RG, MTOR and CIDR are the abbreviations of BLEU, ROUGE, METEOR and CIDEr respectively.}
\centering
\label{tab:main_results}
\begin{tabular}{clccccccc}
\toprule  
\textbf{Dataset} & \textbf{Method} & \textbf{BL-1} & \textbf{BL-2} & \textbf{BL-3} &
\textbf{BL-4}  &\textbf{RG-L} & \textbf{MTOR} &\textbf{CIDR}\\
\midrule

\multirow{9}{*} {\textbf{IU-Xray}}&$ADAATT$ \cite{lu2017knowing} &0.220 &0.127 &0.089 &0.068 &0.308 & - & 0.295 \\
\multirow{9}{*} &$ATT2IN$ \cite{rennie2017self} &0.224 &0.129 &0.089 &0.068 &0.308 & - &0.220 \\
\multirow{9}{*} &$SentSAT+KG$ \cite{zhang2020radiology} &0.441 &0.291 &0.203
&0.147 &0.304 & - &0.304\\
\multirow{9}{*} &$HRGR$ \cite{li2018hybrid} &0.438 &0.298 &0.208 &0.151 &0.322 & - &{0.343}\\
\multirow{9}{*} &$CoAT$ \cite{jing2018automatic} &0.455 &0.288 &0.205 &0.154 &0.369 & - &0.277 \\
\multirow{9}{*} &$CMAS-RL$ \cite{jing2019show} &0.464 &0.301 &0.210 &0.154 &0.362 & - &0.275\\
\multirow{9}{*} &$R2Gen$ \cite{chen2020generating} &0.470 &0.304 &0.219 &0.165 &0.371 &0.187 &-\\ 
\multirow{9}{*} &$KERP$ \cite{li2019knowledge} &0.482 &0.325 &0.226 &0.162 &0.339 & - &0.280\\
\multirow{9}{*} &$CMCL$ \cite{liu-etal-2021-competence} &0.473 &0.305 &0.217 &0.162 &0.378 & 0.186 & -\\
\multirow{9}{*} &$R2GenCMN^{*}$ \cite{chen2021cross} &0.475 &0.309 &0.222
&0.170 &{0.375} & 0.191 &-\\
\multirow{5}{*} &$XPRONet$ \cite{wang2022cross} &\textbf{0.525} &\underline{0.357} &\underline{0.262} &\underline{0.199} &\textbf{0.411} & \textbf{0.220} & \underline{0.359}\\
\cline{2-9}
\cline{2-9}
\multirow{9}{*}  &$\bm{CAMANet} (Ours)$  &\underline{0.504} &\textbf{0.363} &\textbf{0.279} &\textbf{0.218}  &\underline{0.404} & \underline{0.203} &\textbf{0.418}\\
\midrule

\multirow{8}{*}{\textbf{MIMIC}} &$RATCHET$ \cite{hou2021ratchet} &0.232 &- &- &- &0.240 & 0.101 &-\\
\multirow{8}{*}{\textbf{-CXR}} &$ST$ \cite{vaswani2017attention} &0.299 &0.184 &0.121 &0.084 &0.263 & 0.124& - \\
\multirow{8}{*} &$ADAATT$ \cite{lu2017knowing} &0.299 &0.185 &0.124 &0.088 &0.266 & 0.118 &-\\
\multirow{8}{*} &$ATT2IN$ \cite{rennie2017self} &0.325 &0.203 &0.136 &0.096 &0.276 & 0.134 &-\\
\multirow{8}{*} &$TopDown$ \cite{anderson2018bottom} &0.317 &0.195 &0.130 &0.092 &0.267 & 0.128 &- \\

\multirow{9}{*} &$CMCL$ \cite{jing2019show} &0.344 &0.217 &0.140 &0.097 &\textbf{0.281} & 0.133 & -\\

\multirow{9}{*} &$R2Gen$ \cite{chen2020generating} &\underline{0.353} &\underline{0.218} &{0.145} &0.103 &0.277 & \underline{0.142} &0.146\\ 
\multirow{5}{*} &$R2GenCMN$ \cite{chen2021cross} &\underline{0.353} &\underline{0.218} &\underline{0.148} &\underline{0.106} &0.278 & 0.142 &-\\

\multirow{5}{*} &$XPRONet$ \cite{wang2022cross} &0.344 &0.215 &{0.146} &{0.105} &\underline{0.279} & 0.138 &\underline{0.154}\\

\cline{2-9}

\multirow{5}{*}  &$\bm{CAMANet} (Ours)$  &\textbf{0.374} &\textbf{0.230} &\textbf{0.155} &\textbf{0.112} &\underline{0.279} & \textbf{0.145} & \textbf{0.161}\\

\bottomrule 
\end{tabular}
\vspace{-10pt}
\end{table*}

\vspace{-5pt}
\subsection{Implementation Details}
Following the same setting of previous work \cite{li2018hybrid, chen2020generating,chen2021cross}, we utilize both images of a patient on IU-XRay by concatenating the visual tokens, and one view for MIMIC-CXR. Images are firstly resized to $(256, 256)$ and then cropped to $(224, 224)$ (random crop in training and center crop in inference). During the training, we randomly apply one of the operation from $\{rotation, scaling\}$ to further augment the datasets. We employ the DenseNet121~\cite{huang2017densely} pre-trained on ImageNet~\cite{deng2009imagenet} as our visual extractor and a randomly initialized memory driven Transformer~\cite{chen2020generating} as the backbone for the encoder-decoder module with $3$ layers, 8 attention heads and $512$ dimensions for the hidden states. The visual extractor produces $7\times 7$ visual tokens, thus $N^s=49$. 

The Adam~\cite{kingma2015adam} is used to optimize CAMANet. The learning rates are set to $1e-3$ and $2e-3$ for the visual extractor and encoder-decoder on IU-Xray, while MIMIC-CXR has a smaller learning rate with $5e-5$ and $1e-4$ respectively. $\lambda$ and $\delta$ in Equation~\ref{final_loss} are set to $1$ and $0.15$ on IU-Xray, and $1$ and $0.5$ on MIMIC-CXR. The proportion of important words $k$ in \autoref{eq:w_maxpool} is $0.25$ and $0.3$ on IU-Xray and MIMIC-CXR. Note that the optimal hyper-parameters are determined by evaluating the models on the validation sets. 

The same as the most promising studies \cite{li2018hybrid,li2019knowledge,chen2020generating,chen2021cross}, reported, we adopt Beam Search as the sampling method when generating the reports in the validation and test sets. The beam size is set to 3 to balance effectiveness and efficiency. We implement our model via the PyTorch~\cite{paszke2019pytorch} deep learning framework on Nvidia RTX6000 GPU cards. 


\vspace{-5pt}

\subsection{Comparisons to SOTA methods}
We compare CAMANet to previous studies, including the models widely used in conventional image captioning~\cite{lu2017knowing,rennie2017self,anderson2018bottom}, and those proposed for a medical domain~\cite{li2018hybrid,jing2018automatic}. We also compare our method with recent SOTA methods designed for RRG~\cite{jing2019show,chen2020generating,li2019knowledge,chen2021cross,hou2021ratchet,liu-etal-2021-competence,zhang2020radiology,wang2022cross}. They are selected because they share the most similar experimental settings with CAMANet, e.g., the dataset split and the size of the beam search. \textbf{CMAS-RL} takes a two-stage strategy with a cooperative multi-agent system to implicitly capture the abnormalities. \textbf{CMCL} proposed a competence-based multi-modal curriculum learning schema to train the model from easy samples to difficult ones. \textbf{SentSAT + KG} pre-defined a knowledge graph to distill the prior konwledge into the model. \textbf{KERP} decomposed the RRG into explicit medical abnormality graph learning and subsequent natural language modeling. \textbf{R2Gen} designed a relational memory to record key textual information of the generation process. \textbf{R2GenCMN} utilizes a cross-modal memory matrix to record cross-modal patterns to enhance the encoder-decoder framework. \textbf{XPRONet} used a prototype-based cross-modal memory matrix and a multi-label contrastive loss to improve the cross-modal memory matrix learning.

As \autoref{tab:main_results} demonstrates, our proposed method achieves the best performance on all the evaluation metrics MIMIC-CXR datasets except for the RG-L metric on which CAMANet scores slightly lower (-0.2\%) than $CMCL$. Nevertheless, CAMANet surpasses $CMCL$ on all the remaining six evaluation metrics by a notable margin, indicating that CAMANet can generate more semantic reports rather than bias to one indicator such as precision or recall. Similar pattern can be seen on IU-Xray dataset where CAMANet obtains the best scores in four of the seven evaluation metrics and is slightly inferior to XPRONet on the remaining three metrics. Nonetheless, CAMANet surpasses the XPRONet on all the evaluation metrics on the MIMIC-CXR dataset. The possible reason is that XPRONet learns and records the cross-modal prototypes during the training, improving the modelling of informative cross-modal features. Nonetheless, learning the cross-modal prototypes is easier in the small dataset IU-Xray, but is of great difficulty in large dataset such as MIMIC-CXR which is almost 115 times the size of IU-Xray. This further demonstrates that CAMANet possess better generalization capability.

Moreover, CAMANet outperforms the previous SOTA approaches by a notable margin. For instance, CAMANet surpasses the second-best BLEU-4 and CIDEr scores by 1.9\% and 5.9\% respectively on the IU-Xray dataset. Similarly, a 2.1\% and 0.6\% improvements can be seen on BLEU-1 and BLEU-4 scores of CAMANet compared to the second-best performing method on MIMIC-CXR. Note that the MIMIC-CXR dataset is much larger and challenging 
and making even a small improvement proves difficult. 

We mainly attribute the superiority of CAMANet to the improved cross-modal alignment achieved by the VTAC module, and the enriched disease-related feature representations by the proposed VDM assisted encoder. Further visualizations are provided in Section~\ref{sec:qualitative_results}. The following ablation studies confirm the effectiveness of each component in CAMANet. 

\noindent \textbf{Use of Pseudo Labels} CAMANet utilizes the pseudo labels to generate the CAMs. Note that the visual extractor of CAMANET is IMAGENET pre-trained. Methods $SentSAT+KG$~\cite{zhang2020radiology} and $RATCHET$~\cite{hou2021ratchet} utilize pseudo labels to pre-train the visual extractor; HRGR adopts two auxiliary datasets with ground truth labels to pre-train the visual extractor; whereas KERP utilizes some manually labelled data and templates; $CMCL$~\cite{liu-etal-2021-competence} uses the pseudo labels to fine-tune their visual extractor and determines the visual difficulty; XPRONet~\cite{wang2022cross} utilizes the pseudo labels to form a cross-modal prototype-driven network. CAMANet still outperforms these methods. Furthermore, the pseudo-labels we used are provided with the MIMIC-CXR dataset.


\vspace{-5pt}
\subsection{Clinical Efficacy}
  To verify that our model can better capture the abnormal information, we further compare the clinical efficacy of CAMANet with recent SOTA RRG methods. Note that clinical efficacy metrics only apply to MIMIC-CXR because the labeling schema of CheXbert is designed for MIMIC-CXR. As can be seen, CAMANet outperforms the previous methods by a large margin on the precision and obtains the best result on the F1-score. In addition, CAMANet is also the second best-performing method by recall where the score is slight lower (-0.02) than the R2GenCMN. Note that R2GenCMN utilizes a large cross-modal trainable memory matrix ($2048 \times 512$) which is expanded to each sample during the training to record the cross-modal patterns, significantly increasing the GPU memory requirement. Nonetheless, CAMANet still shows obviously better results on precision and F1-score with slight increase for the training cost.

\begin{table}[ht]
\caption{Clinical efficacy comparisons on MIMIC-CXR. $\star$ means that results are obtained by using their provided trained models and released official code to generate the reports.}
\centering
\label{tab:main_results}
\begin{tabular}{l|cccc}
\toprule  
\textbf{Methods}  & \textbf{Precision} & \textbf{Recall} & \textbf{F1-Score}   & 
\\
\midrule  

$R2Gen^\star$   &0.406 &0.213 &0.280  \\
$R2GenCMN^\star$  &0.440 &\textbf{0.325} &\underline{0.374} \\
$XPRONet^\star$ &\underline{0.463} &0.285 &0.353  \\
CAMANet  &\textbf{0.483} &\underline{0.323} &\textbf{0.387} \\
\bottomrule 
\end{tabular}
\vspace{-5pt}
\end{table}

\subsection{Comparisons to PLMs and PVLM}
We further compare our models with one recent approach \cite{wang2023can} leveraging PLMs with prompt learning, and four recent PVLMs including two in medical/radiology domain \cite{moon2022multi, delbrouck2022vilmedic} and two in traditional scene domain \cite{dou2022empirical,dou2022coarse}. As can be seen in Table 2, CAMANet achieves competitive results to PromptRRG where CAMANet obtains better or same scores on three out of five evaluation metrics. Note that PromptRRG replaces the embedding layer with PLMs and leverages prompt learning to distill the prior knowledge into the entire model. Although demonstrating improvements in image retrieval and visual-question answering, the pre-trained vision-language models listed fails to show promising results in RRG possibly since the simple pre-training scheme cannot cope with the intrinsic problems in RRG aforementioned and inadequate pre-training paired strong annotation medical data for a cross-modal task such as RRG. Another reason is that the commonly used pre-training objectives, e.g., image-text matching, could be more difficult to take effect in RRG (see analysis of CLIP in ablation study).




\begin{table}
\caption{Comparative results of CAMANet with some recent studies leveraging the PLMs and four PVLMs on MIMIC-CXR dataset.}
\centering
\label{tab:main_results}
\begin{tabular}{l|ccccc}
\toprule  
\textbf{Methods}  & \textbf{BL1} & \textbf{BL4} &\textbf{RG-L} &\textbf{MTOR} & \textbf{CIDEr} \\
\midrule  

$PromptRRG$~\cite{wang2023can} &0.348 &\textbf{0.113} &0.263 &\textbf{0.145} &\textbf{0.286} \\
$MedViLL$ \cite{moon2022multi} &-  &0.066 &- &- &- \\
$ViLMedic$ \cite{delbrouck2022vilmedic} &- &0.082 &0.225 &- &- \\
$METER$ \cite{dou2022empirical} &0.308 &0.088 &0.232 &0.122 & 0.179 \\
$FIBER$ \cite{dou2022coarse} &0.307 &0.093 &0.240 &0.123 & 0.128 \\
\hline
$CAMANet$ &\textbf{0.374} &\underline{0.112} &\textbf{0.279} & \textbf{0.145} &\underline{0.161}  \\

\bottomrule 
\end{tabular}
\vspace{-10pt}
\end{table}

\begin{table*}
\caption{The experimental results of ablation studies on the IU-Xray and MIMIC-CXR datasets.}
\centering
\label{tab:ablation_studies}
\begin{tabular}{l|cccccccc}
\toprule  
\textbf{IU-Xray}  & \textbf{BL-1} & \textbf{BL-2} & \textbf{BL-3} &
\textbf{BL-4}  &\textbf{RG-L} & \textbf{MTOR} & \textbf{CIDR} & \textbf{AVG.$\Delta$} \\
\midrule  

Base  &0.451 &0.289 &0.209 &0.159 &0.365 & 0.175 &0.336 &- \\

+VDMAE  &0.473 &0.321 &0.238 &0.177 &0.402 & 0.189 &0.415 &11.8\%\\
+VDMAE+VTAC &\textbf{0.504} &\textbf{0.363} &\textbf{0.279} &\textbf{0.218} &\textbf{0.407} &\textbf{0.203} &\textbf{0.418} &\textbf{22.8\%}\\

\toprule  
\textbf{MIMIC-CXR}  & \textbf{BL-1} & \textbf{BL-2} & \textbf{BL-3} &
\textbf{BL-4}  &\textbf{RG-L} & \textbf{MTOR} & \textbf{CIDR} &\textbf{AVG.$\Delta$} \\
\midrule  

Base  &0.324 &0.203 &0.138 &0.100 &0.276 & 0.135 &0.144 &-\\
+VDMAE  &0.357 &0.227 &0.148 &0.107 &0.277 & 0.142 &\textbf{0.162} & 7.7\% \\
+VDMAE+VTAC  &\textbf{0.374} &\textbf{0.230} &\textbf{0.155} &\textbf{0.112} &\textbf{0.279} & \textbf{0.145} &0.161 &\textbf{10.5\%} \\ 
\hline
+VDMAE+CLIP(0.1)  &0.342 &0.211 &0.142 &0.102 &0.278 &0.138 &0.154 &\textbf{3.5\%} \\

+VDMAE+CLIP(0.25)  &0.331 &0.204 &0.137 &0.098 &0.272 &0.134 &0.145 &\textbf{0.0\%} \\

\bottomrule 
\end{tabular}
\end{table*}

\vspace{-5pt}
\subsection{Ablation Studies}
Here, we firstly explore the influence of each component in CAMANet including the VDMAE and 
VTAC module. The following models are used to conduct the ablation studies:
\begin{itemize}[noitemsep,nolistsep]
\item \textbf{Base}: The base model consists only of the visual extractor and the base encoder-decoder without other extensions.
\item \textbf{Base+VDMAE}: The standard encoder in the base model is replicated with our proposed VDM assisted encoder by adding the VDM Generation and VDMAE module. 
\item \textbf{Base+VDMAE+VTAC}: This is the full CAMANet containing all our proposed components.
\end{itemize}
We present the main ablation study results in \autoref{tab:ablation_studies}. 
A remarkable improvement can be seen by adding the VDMAE module, e.g., CIDEr scores increase from $0.336$ to $0.415$, and $0.144$ to $0.162$ on IU-Xray and MIMIC-CXR benchmarks respectively. Our method gains the most significant improvements when further integrating with the VTAC module. In particular, the full model obtains an average of $\bm{22.8}$\% improvement over all the evaluation metrics on IU-Xray, and $\bm{10.5}$\% on MIMIC-CXR. The improvements on MIMIC-CXR are less obvious than IU-Xray since it is harder to learn a robust discriminative representation and cross-modal alignments in such a large dataset (50 times larger than IU-Xray). Section \ref{sec:qualitative_results} presents some qualitative results to further illustrate the model's effectiveness.

To further demonstrate the effectiveness of our VTAC module, we compare it with a recent, widely-used image-text alignment technique CLIP~\cite{radford2021learning} on the MIMIC-CXR dataset. Note that CLIP focuses on the image-text alignment while CAMANet is designed for a more fine-grained level, i.e., region-word alignments. Specifically, we adapt the CLIP technique to our base model by taking the averaged visual tokens and the averaged textual tokens from the last decoder layer to form an image-text matching task. We verify two set of CLIP loss weights, i.e., $0.1$ and $0.25$ and demonstrate the comparison in the last two rows in \autoref{tab:ablation_studies}. As can be seen, adding the CLIP loss significantly harms the performance compared with its base model (Base+VDMAE) and reducing the contribution of the CLIP loss reduces the negative influence. The reason could be that images from different patients could have the same or almost the same reports especially for the normal cases in RRG. Furthermore, even in pathological cases, most report sentences may be associated with a description of normal findings. Therefore, image-text level alignment is more difficult to take effect in RRG and may confuse the model given normally unsatisfactory visual feature extractors in RRG. Nonetheless, our proposed VTAC module focuses on region-word alignments with each report, hence suffering less from this problem and demonstrating improved performance.

\vspace{-10pt}

\subsection{Sensitivity of the Proportion of Selected Words}
\label{sec:k_ablation}
\begin{figure}
    \centering 
    \includegraphics[width=0.4\textwidth]{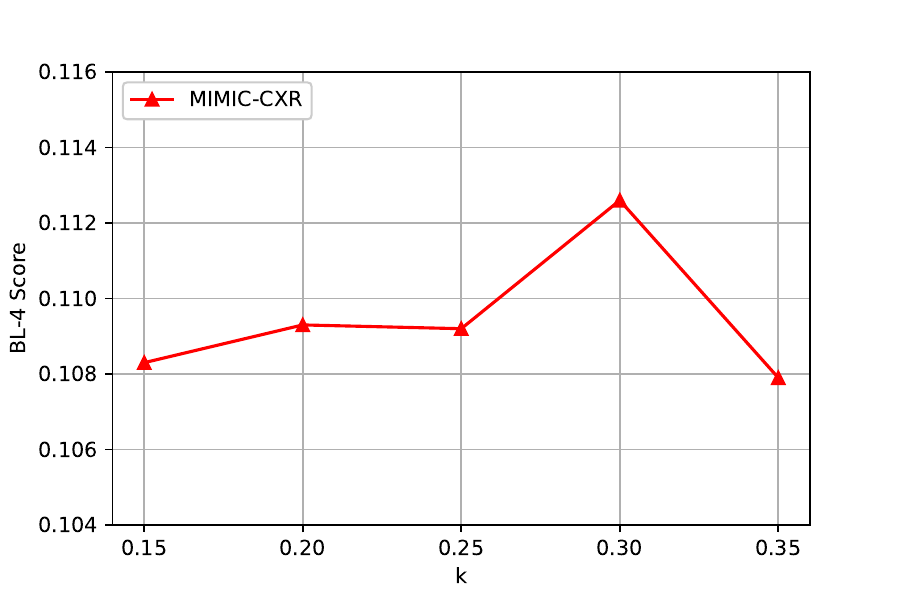} 
    \caption{Effect of varying $k$ 
    on MIMIC-CXR. (BLEU-4 score).}    
    \label{fig:k_ablation_mimic}  
    \vspace{-10pt}
\end{figure}

\begin{figure}
    \centering 
    \includegraphics[width=0.4\textwidth]{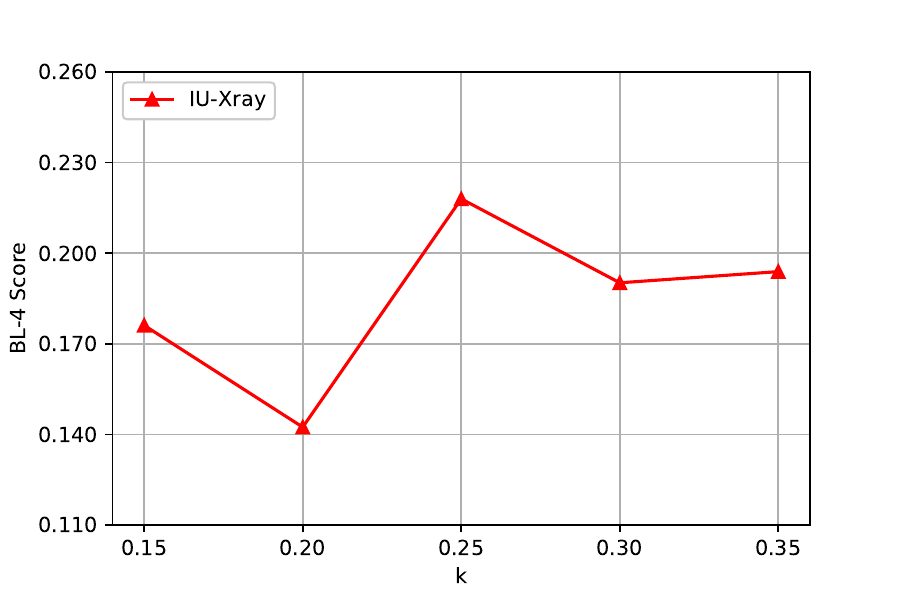} 
    \caption{Effect of varying $k$ 
    on IU-Xray. (BLEU-4 score).}    
    \label{fig:k_ablation_iu}  
    \vspace{-15pt}
\end{figure}

To investigate the sensitivity of the hyper-parameter $k$, we vary the proportion of the selected important word tokens from $0.15$ to $0.35$. As illustrated in \autoref{fig:k_ablation_mimic} and \autoref{fig:k_ablation_iu}, CAMANet is not overly sensitive to this proportion.  However, it is still important to strike a balance when setting the value of $k$ as a small $k$ may not encompass all the important words, while too large a value may introduce irrelevant words. For example, a performance drop can be seen on the IU-Xray dataset when $k$ increases from 15 to 20. This might be caused by the introduction of unimportant (noisy) words such as ``there" and ``are". Further increasing $k$ to 25 brings a notable improvement, demonstrating that the newly introduced words bring more benefits and offset any negative influence from previously noisy words. The fluctuation on IU-Xray is relatively greater than MIMIC-CXR, which is expected since IU-Xray is significantly smaller and the reports contain larger word variance compared to MIMIC-CXR.

\begin{figure*}
    \centering 
    \includegraphics[width=0.95\textwidth]{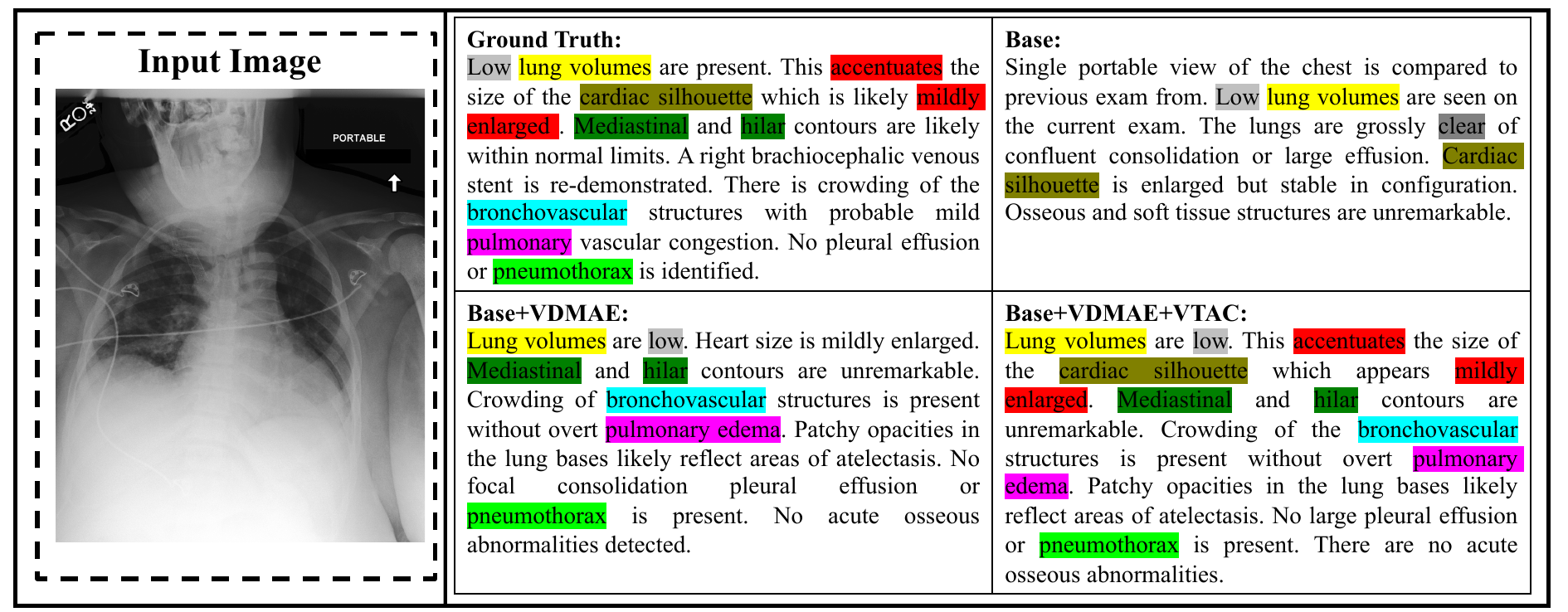} 
    \caption{An example of the reports generated by different models. Most medical terms are highlighted 
    to better differentiate the findings of report.}    
    \label{fig:example_vis}  
    \vspace{-15pt}
\end{figure*}

\begin{figure}
    \centering 
    \includegraphics[width=0.4\textwidth]{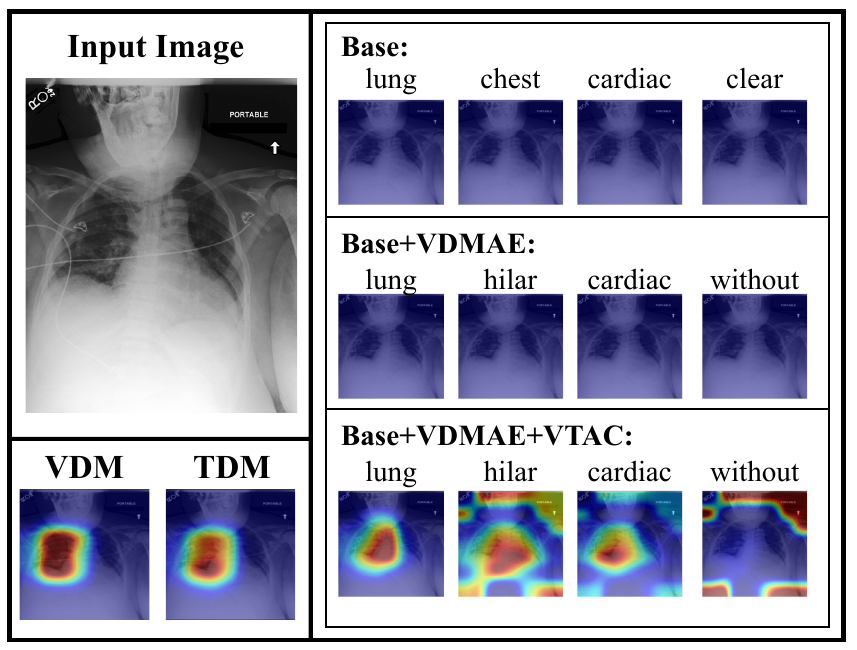} 
    \caption{The visualizations of cross-modal attention maps for some medical terms by different models. The VDM and TDM are also shown and appear consistent with the GT report because: (1) they correctly highlight the right-lung; (2) the TDM hot-spots focus on the probable mild pulmonary vascular congestion noted by the radiologist.}
    \label{fig:attn_vis}  
    \vspace{-5pt}
\end{figure}

\subsection{Qualitative Results}
\label{sec:qualitative_results}
To further verify the proposed method, we provide an example in \autoref{fig:example_vis} which shows the reports generated by different models. CAMANet appears to better capture the disease information and generate the abnormal description, e.g., \textit{``...This accentuates the size of the cardiac silhouette which appears mildly enlarged...''}, compared to the base model. Moreover, we visualize its generated visual discriminative map (VDM), textual discriminative map (TDM) and the cross-modal attention map from the last decoder in \autoref{fig:attn_vis} to explore whether the cross-modal alignment is truly enhanced by the proposed VTAC module. It can be seen that the \textit{Base} model cannot learn any useful cross-modal alignments when generating the reports. Similar to the \textit{Base} model, \textit{Base+VDMAE} also fails in the cross-modal attention. This is expected since the discriminative representation is detached from the decoder. Nonetheless, owing to the enriched disease information, the \textit{Base+VDMAE} model still can capture some abnormal single-model patterns, and therefore achieves improved performance compared to the \textit{Base} model. It is clear that after adding the VTAC module, CAMANet demonstrates potent cross-modal alignment capability where the medical terms, e.g., `\emph{lung}', focus on the associated discriminative image regions, while trivial words, e.g., `\emph{without}', pay attention only to the background. Visualizations of the VDM and TDM further verify the effectiveness of employing a VDM as a form of pseudo label to supervise the cross-modal attention learning.

\begin{figure*}[!h]
    \centering 
    \includegraphics[width=0.92\textwidth]{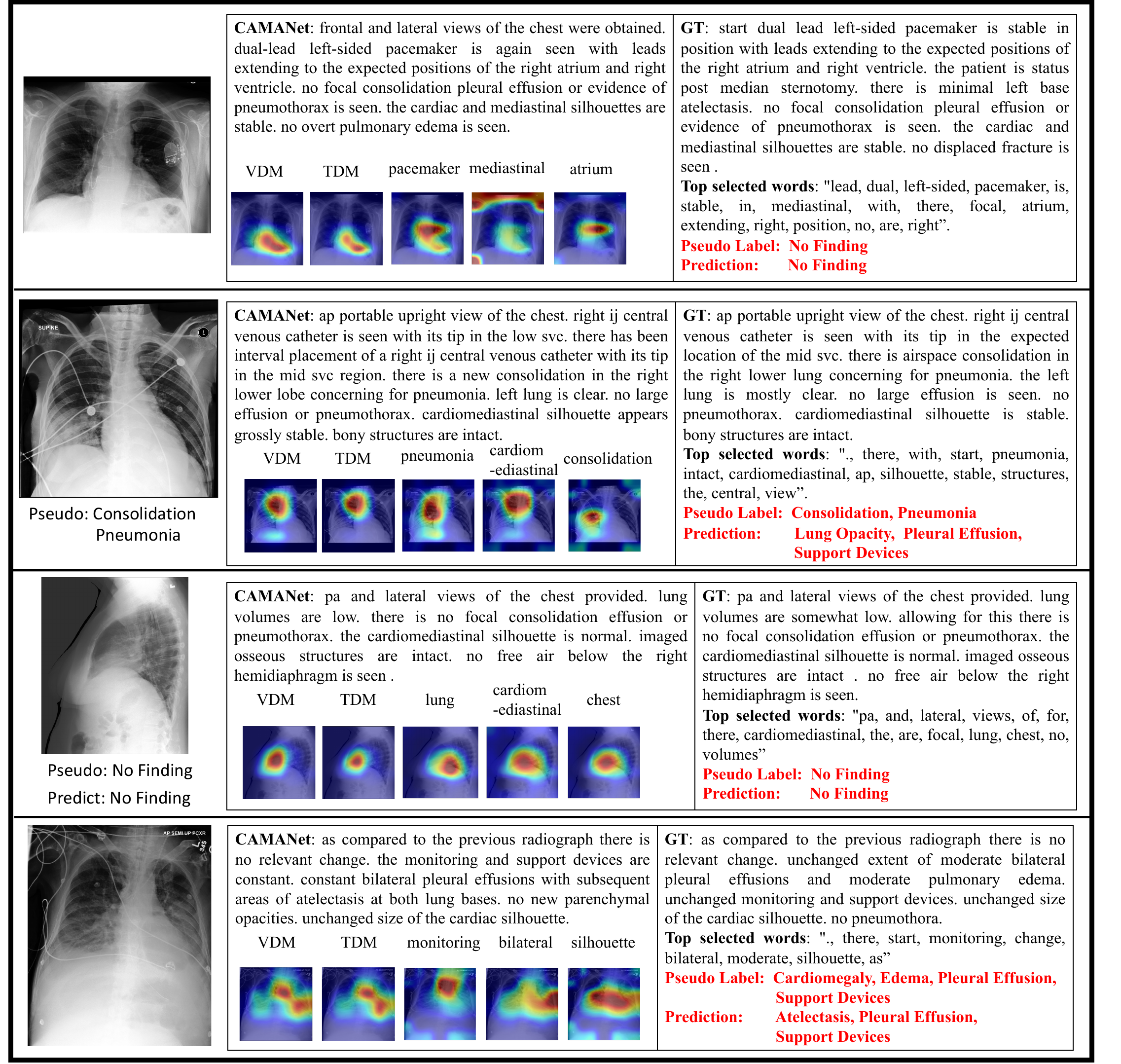} 
    \caption{Some examples of generated reports by CAMANet and the associated VDM, TDM, Cross-modal attention maps and selected important words. Duplicated words in the selected important words have been removed.}    
    \label{fig:more_examples}  
    \vspace{-15pt}
\end{figure*}

Finally we present further visualization results in~\autoref{fig:more_examples}. The visualizations show the reports predicted by CAMANet and the associated ground truth. We also provide the visual discriminative map (VDM), textual discriminative map (TDM), the cross-attention maps of some medical terms in the inference mode and include important words in the training.

\vspace{-5pt}
\subsection{Multi-label Classification Performance}
To further analyze whether the visual extractor can truly provide a useful discriminative map, we also evaluate the classification performance of the visual extractor on the validation and test sets of IU-Xray and MIMIC-CXR using precision and recall metrics. As \autoref{tab:mp} shows, CAMANet achieves encouraging results on both sets of the IU-Xray which may partially explain the remarkable improvements of CAMANet on IU-Xray. Nonetheless, on MIMIC-CXR, CAMANet shows reasonably good results in the validation set, while performing worse in the test set with a precision score of 0.6747 and a recall score of 0.3871. We summarize the possible reasons below: (1) The IU-Xray dataset provides both the frontal and lateral view of the radiology images for each patient while MIMIC-CXR only has one view (frontal or lateral), making image classification more challenging on MIMIC-CXR; (2) the visual tokens are mainly tailored for the main task of report generation, rather than image classification; (3) Noisy and inaccurate pseudo-labels in a large dataset such as MIMIC-CXR have a more detrimental effect compared to smaller datasets. 
Nevertheless, an imperfect visual extractor can still bring notable improvements for report generation on MIMIC-CXR. During the experiments, we found that even if a sample is wrongly classified, the model may still produce an activation map focusing on the foreground, e.g, chest or lung, rather than the background, reducing the reliance on pseudo-labels. The second sample in \autoref{fig:more_examples} demonstrates that a sample with pseudo label ``\textit{Consolidation, Pneumonia}" is wrongly classified as “\textit{Lung Opacity, Pleural Effusion, Support Devices}”. Therefore, the model fails to pay attention to the specific, fine-grained abnormal regions. However, the model can still generate a visual discriminative map and attention maps (see the attention maps of word ``\textit{pneumonia}" and ``\textit{consolidation}") focusing more on a relative coarse-grained ``\textit{lung}” region since the wrong prediction still shares some semantic information as pseudo-labels, i.e., lung-related diseases. In addition, the wrong pseudo label of “\textit{Lung Opacity, Pleural Effusion, Support Devices}” contains the word of “\textit{Effusion}”, which appeared in the ground truth report, "\textit{no large effusion is seen}". This example shows that the activation map derived from a wrongly classified sample could still be useful in directing the model attention to the image region corresponding to the relevant organ, and the description of the pseudo label may find overlapping words in the ground truth report.

\begin{table}
\begin{center}
\begin{tabular}{ccc|ccc}
\hline
\multicolumn{3}{c|}{IU-Xray} &\multicolumn{3}{c}{MIMIC-CXR} \\ 
\toprule
Subset &Precision & Recall &Subset &Precision & Recall \\
\hline
Val & 0.7551 &0.7639 &Val  & 0.7255 &0.5413   \\
\hline
Test & 0.8000 & 0.7915 &Test & 0.6747 &0.3871   \\
\hline
\end{tabular}
\end{center}
\caption{The multi-label classification results on the MIMIC-CXR and IU-Xray. Val and Test refer to the validation and test set.}
\label{tab:mp}
\vspace{-15pt}
\end{table}

\vspace{-5pt}
\subsection{Computational Costs}
\label{sec:training_time}

\paragraph{Training and Testing Time}
The batch size is set to 32 for IU-Xray and 64 for MIMIC-CXR. We train all the models for 30 epochs. 
\autoref{tab:tt} shows the average training and testing time of one epoch for the \textit{Base} and \textit{CAMANet} on IU-Xray and MIMIC-CXR. CAMANet only slightly increases the training/testing time on MIMIC-CXR, while the increased computational cost is negligible on IU-Xray.
\begin{table}
\begin{center}
\begin{tabular}{|c|cc|cc|}
\hline
\multirow{2}{*}{Dataset} &\multicolumn{2}{c|}{Train} &\multicolumn{2}{c|}{Test}\\ \cline{2-5}
&Base & CAMANet &Base &CAMANet \\
\hline
IU-Xray &0.41  &0.42 &1.28 &1.32   \\
\hline
MIMIC &56.35  &60.06  &13.27 &13.46 \\
\hline
\end{tabular}
\end{center}
\caption{The average training/test time (minutes) of one epoch for different models on the training/test set.}
\label{tab:tt}
\vspace{-15pt}
\end{table}

\paragraph{The number of parameters}
\label{sec:model_parameters}
Here, we compare the number of parameters of the \textit{Base} and \textit{CAMANet} models in \autoref{tab:mp}. The increase in number of learnable parameters is negligible compared to the base model.
\begin{table}
\begin{center}
\begin{tabular}{|c|c|c|}
\hline
 Method &Base & CAMANet \\
\hline
\#Param &46.07M &46.09M   \\
\hline
\end{tabular}
\end{center}
\caption{The number of parameters of different models.}
\label{tab:mp}
\vspace{-20pt}
\end{table}

\vspace{-5pt}
\subsection{Limitations and Future Works}

Although the accuracy requirement of the pseudo labels is not critical, CAMANet still needs pseudo labels to develop the VDMAE and VTAC modules. We believe that CAMANet could bring greater improvements when given more accurate pseudo-labels. In addition, selected important words sometimes do not seem to capture the most important medical terms or, such important medical terms are not ranked at the top, possibly introducing noise in to the model. One possible reason is that the selection of the important words rely on the learning process which is based on the similarities between the discriminative representation and the word embeddings, while the learning of the discriminative representation is imperfect due to the inaccurate pseudo labels and influence from the main task, i.e., report generation (visual tokens are mainly tailored for main task, i.e., report generation, rather than image categorization.). Nonetheless, the inclusion of these commonly used words could also be reasonable in some cases as they may share the semantic information with these medical terms. For example, for the sentence ``\textit{there is pleural effusion}”, since the model generates the word one by one, words ``\textit{there is}” here will indicate the presentation of an important medical term. 

The failure cases include: (1) samples with very inaccurate pseudo labels, e.g., heart diseases to be labelled as lung diseases; (2) some images with rare descriptions fail to generate the abnormal findings as the severe data bias problem is not completely solved; (3) some images fails to generate the normal description such as ``\textit{bony structures are intact}". However, we think some failures are also reasonable. For example, the radiologists' behaviours towards the inclusion of some normal descriptions, e.g, ``\textit{bony structures are intact}", are different as it seems acceptable to not mention the ``\textit{bony structures are intact}" if no abnormal findings are detected. This, however, could confuse the model. 

One future work could be how to design a more accurate method for selecting important words. In addition, alleviating the negative influence of noisy pseudo-labels plays an essential role in further improving the performance. 
Moreover, we utilize vanilla CAM in this work and more sophisticated CAM methods are highly expected to bring more promising results.

\vspace{-5pt}

\section{Conclusions}

In this work, we propose a novel class activation map guided RRG framework, CAMANet, which explicitly leverages cross-modal alignment and disease-related representation learning. Our visual discriminative map assisted encoder distills the discriminative information into the model via a derived discriminative representation and the self-attention mechanism. The generated VDM is then regarded as the ground truth in the proposed visual-textual attention consistency module to supervise the cross-modal attention learning, aimed at explicitly promoting cross-modal alignment. Experimental results on two widely used RRG benchmarks prove the superiority of CAMANet over previous studies. The ablation studies further verify the effectiveness of individual components of CAMANet. Moreover, a number of illustrative visualizations and discussions are provided to inspire future research. 

\vspace{-5pt}
\section*{Declaration of competing interest}
The authors declare that they have no known competing financial interests or personal relationships that could have appeared to influence the work reported in this paper.


\bibliographystyle{IEEEtran}
\bibliography{refs}

\end{document}